\documentclass[letterpaper, 10 pt, conference]{ieeeconf}
\IEEEoverridecommandlockouts
\usepackage[utf8]{inputenc}
\usepackage[noend]{algpseudocode}
\usepackage{amsmath}
\usepackage{amssymb}    
\usepackage[utf8]{inputenc}
\usepackage{xcolor}
\usepackage{graphicx}
\usepackage{hyperref}
\usepackage{booktabs}
\usepackage{color}
\usepackage{array}
\usepackage{booktabs}
\usepackage{multirow}
\usepackage{geometry}
\usepackage{hyperref}
\usepackage[space]{cite}
\usepackage[linesnumbered,boxed,ruled,commentsnumbered]{algorithm2e}
\graphicspath{{figures/}}
\DeclareGraphicsExtensions{.pdf,.png,.jpg,.eps}
\begin{document}
	%
	\title{\LARGE \bf FAST-LIVO: Fast and Tightly-coupled Sparse-Direct  LiDAR-Inertial-Visual Odometry}

	\author{Chunran Zheng$^{1*}$,
		Qingyan Zhu$^{1*}$,
		Wei Xu$^{1}$,
		Xiyuan Liu$^{1}$,
		Qizhi Guo$^{1}$
		and Fu Zhang$^{1}$,
		\thanks{$^*$These two authors contribute equally to this work.}
		\thanks{$^{1}$C. Zheng, Q. Zhu, W. Xu, X. Liu, Q. Guo and F. Zhang are with the Department of Mechanical Engineering, The University of Hong Kong, Hong Kong Special Administrative Region, People's Republic of China.
			{\tt\footnotesize $\{$zhengcr,xuweii,xliuaa$\}$@connect.hku.hk}, {\tt\footnotesize $ $fuzhang$ $@hku.hk}
		}
	}
	
	\markboth{Journal of \LaTeX\ Class Files,~Vol.~6, No.~1, January~2007}%
	{Shell \MakeLowercase{\textit{et al.}}: Bare Demo of IEEEtran.cls for Journals}
	
	\maketitle
	\thispagestyle{empty}
	
	\begin{abstract}
		\textcolor[rgb]{0,0,0}{To achieve accurate and robust pose estimation in Simultaneous Localization and Mapping (SLAM) task, multi-sensor fusion is proven to be an effective solution and thus provides great potential in robotic applications. This paper proposes FAST-LIVO, a fast LiDAR-Inertial-Visual Odometry system, which builds on two tightly-coupled and direct odometry subsystems: a VIO subsystem and a LIO subsystem. The LIO subsystem registers raw points (instead of feature points on e.g., edges or planes) of a new scan to an incrementally-built point cloud map. The map points are additionally attached with image patches, which are then used in the VIO subsystem to align a new image by minimizing the direct photometric errors without extracting any visual features (e.g., ORB or FAST corner features). To further improve the VIO robustness and accuracy, a novel outlier rejection method is proposed to reject unstable map points that lie on edges or are occluded in the image view.  Experiments on both open data sequences and our customized device data are conducted. The results show our proposed system outperforms other counterparts and can handle challenging environments at reduced computation cost. The system supports both multi-line spinning LiDARs and emerging solid-state LiDARs with completely different scanning patterns, and can run in real-time on both Intel and ARM processors. We open source our code and dataset of this work on Github\footnote[2]{\url{https://github.com/hku-mars/FAST-LIVO}} to benefit the robotics community.}
		
	\end{abstract}
	
	\IEEEpeerreviewmaketitle
	
	\section{Introduction}
	In recent years, simultaneous localization and mapping (SLAM) has made great progress in real-time 3D reconstruction and localization in unknown environments. Currently, there are several successful implemented framework using a single measurement sensor such as camera \cite{mur2015orb,mur2017orb} or LiDAR \cite{zhang2014loam, lin2020loam, shan2018lego}. However, with the increasing demand of operating intelligent robots in real world that usually contains challenging structure-less or texture-less environments, existing systems using a single sensor cannot achieve an accurate and robust pose estimation as required. To address this issue, multi-sensor fusion \cite{qin2018vins, qin2019lins, hyun2020uwb, kramer2020radar} has drawn increasing recent attention to combine the advantages of different sensors and provide an effective pose estimation in sensor degraded environments, showing great potentials in robotic applications.
	
	Among those sensors used in robotics, camera, LiDAR and Inertial Measurement Units (IMUs) are possibly the mostly widely used sensors in SLAM tasks. Several recent LiDAR-inertial-visual odometry systems (LIVO) have been proposed to achieve robust state estimation such as R2LIVE \cite{lin2021r2live} and LVI-SAM \cite{shan2021lvi}. They usually include a LiDAR-inerial odometry (LIO) subsystem and a visual-inerial odometry (VIO) subsystem that jointly fuses the state vector but separately process each data without considering their measurement-level coupling. The resultant system usually takes up significant computation resources. To address this issue, we propose a fast and tightly-coupled sparse-direct LiDAR-inertial-visual odometry system (FAST-LIVO), combining the advantage of sparse direct image align with direct raw points registration to achieve accurate and reliable pose estimation at a reduced computation cost. The contributions of this paper are listed as below:
	\begin{enumerate}
		
		\item A compact LiDAR-inertial-visual odometry framework, which builds on two direct and tightly-coupled odometry systems: a LIO subsystem and a VIO subsystem. These two subsystems estimates the system state jointly by fusing their respective LiDAR or visual data with IMUs. 
		
		\item A direct and efficient VIO subsystem that maximally re-use the point cloud map built in LIO subsystem. Specifically, points in the map are attached with image patches previously observed and then projected to a new image to align its pose (hence full system state), by minimizing the direct photometric errors. The LiDAR points re-use in VIO subsystem avoids the extraction, triangulation or optimization of visual features and couples the two sensors at the measurement level.
		
		\item Implementation of the proposed system into a practical open software that can run in real-time on both Intel and ARM processors, and supports both multi-line spinning LiDARs and emerging solid-state LiDARs with completely different scanning patterns. 
		
		\item Verification of the developed system on both open data sequences (i.e., NTU VIRAL Dataset \cite{nguyen2021ntuviral}) and our customized device data. Results show that our system outperforms other counterparts and can handle challenging sensor-degenerated environments at a reduced computation cost.
		
	\end{enumerate}
	
	\section{Related works}
	Previous works related to our system and the involved techniques can be divided into the following two parts.
	
	\subsection{Direct Methods\label{direct-method}}
	Direct method is one of the most popular ways to achieve fast pose estimation in both visual and LiDAR SLAM. Different from feature-based methods (\cite{qin2018vins, mur2015orb, zhang2014loam, lin2019loam_livox}) that require to extract salient feature points (e.g., corners and edge pixels in images; plane and edge points in LiDAR scans), and generate robust feature descriptors for correspondence matching. Direct methods directly use raw measurements to optimize sensor pose \cite{driectmethod-1} by minimizing an error function using photometric error or point-to-plane residuals, e.g., \cite{forster2014svo}, \cite{9697912}. This can achieve fast pose estimation by eliminating the feature extraction and matching that are usually time-consuming. Due to the lack of robust features matching, these direct methods are highly dependent on good state initialization. Dense direct method is mostly used in tracking for RGB-D cameras like \cite{meilland2011real,tykkala2011direct} and \cite{kerl2013robust}, which includes a dense depth-map in each image frame, and apply image-to-model alignment for estimation. Sparse direct method \cite{forster2014svo} shows a robust and accurate state estimation using only a few well-selected raw patches, thus further reduce the computation load compared by dense direct method. 
	
	Our work maintains the idea of direct method in both LIO and VIO. The LIO of our system is directly adapted from FAST-LIO2 \cite{9697912}. The VIO subsystem is based on sparse direct image align similar to \cite{forster2014svo} but re-uses the map points built in the LIO subsystem to save the time-consuming backend (i.e., feature alignment, sliding window optimization and/or depth filtering).  
	
	\subsection{LiDAR-Visual-Inertial SLAM\label{LiDAR-Visual-Inertial SLAM}}	
	The multiple sensors used in LiDAR-visual-inertial SLAM make it possible to handle various of challenging environments where one sensor fails or is partially degenerated. Motivated by this, there have been several LiDAR-visual inertial SLAM systems developed in the community. To name a few, LIMO \cite{graeter2018limo} extracts the depth information from LiDAR measurements for image features and estimates motion among keyframes based on bundle adjustment. R2LIVE \cite{lin2021r2live} is a tightly-coupled system to fuse the LiDAR-visual-inertial sensors, which achieves an accurate state estimation by extracting LiDAR and image features and then conducts re-projection error within the framework of Iterated Error State Kalman Filter \cite{bell1993iekf}. For the part of VIO subsystem, a sliding window optimization is used to further improve the accuracy of visual features in the map.  LVI-SAM \cite{shan2021lvi} is a feature-based framework similar to \cite{lin2021r2live} but also incorporates loop closure. These feature-based methods are usually computationally-costly due to the extraction of image and/or LiDAR feature points and the subsequent sliding window optimization. Compared with these methods, our proposed FAST-LIVO uses raw LiDAR points and image pixels to respectively track LiDAR scans and images without extracting any features, resulting in higher computation efficiency. 
	
	The VIO subsystem of our FAST-LIVO is most similar to DVL-SLAM \cite{shin2018direct, shin2020dvl}, which projects LiDAR points into a new image and tracks the image by minimizing the direct photometric error. However, DVL-SLAM conducts only frame-to-frame image alignment and does not incorporate any IMU measurements nor LiDAR scan registration. In contrast, our system FAST-LIVO tightly couples the frame-to-map image alignment, LiDAR scan registration, and IMU measurements in an integrated Kalman filter. 
	We also propose a reliable and robust outlier rejection solution when projecting LiDAR measurements to image planes, which further improves the system accuracy.
	
	\section{System overview}
\begin{figure}[t]
		\centering
		\includegraphics[width=1.0\columnwidth]{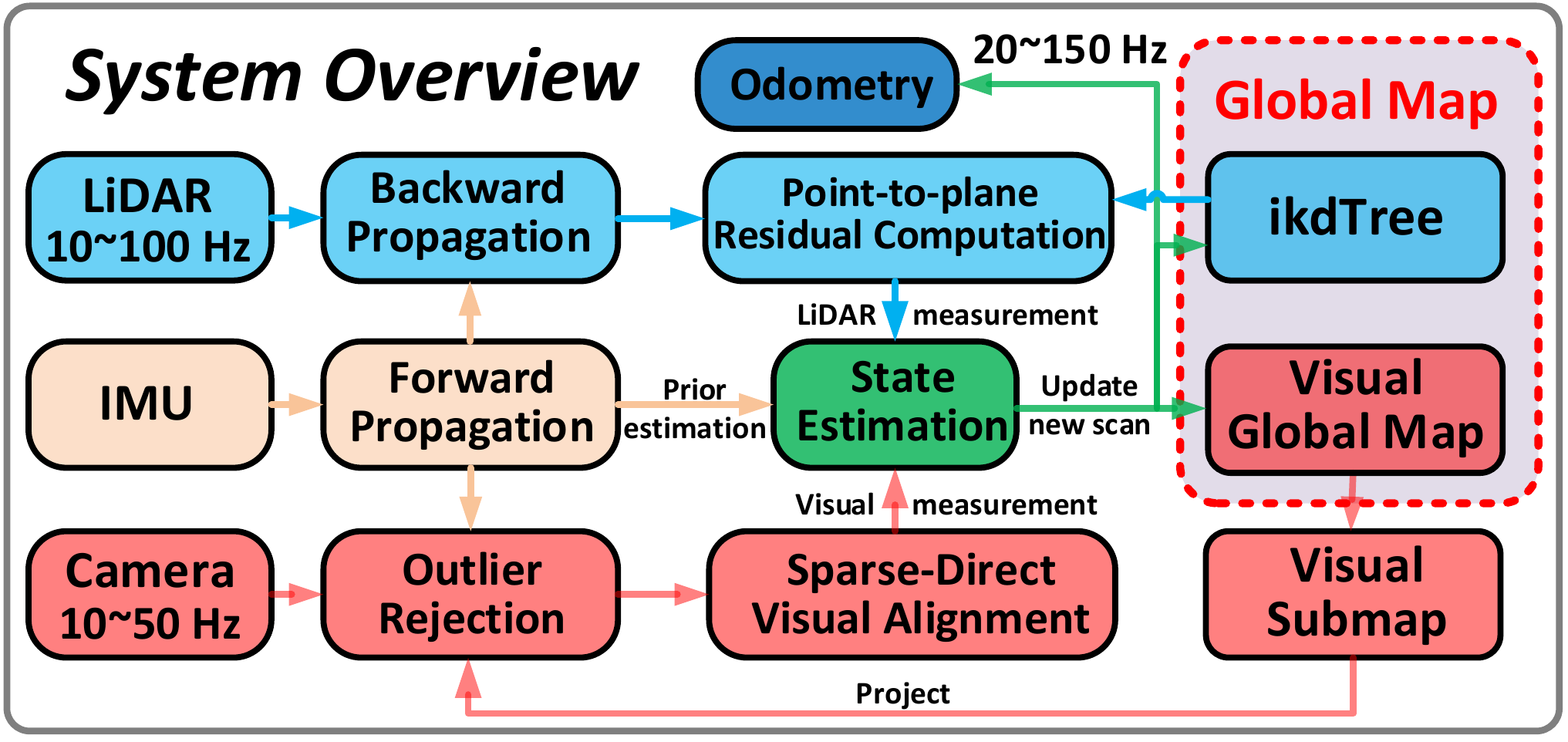}
		\caption{System overview of FAST-LIVO}
		\label{framework}
		\vspace{0cm}
	\end{figure}

	This paper uses notations in Table \ref{tab:symbols}. The overview of our system is shown in Fig. \ref{framework}, which contains two subsystems: the LIO subsystem (the blue part) and the VIO subsystem (the red part). The LIO subsystem first compensates the motion distortion in a LiDAR scan by backward propagation \cite{xu2020fastlio} and then computes the frame-to-map point-to-plane residual. Similarly, the VIO subsystem extracts the visual submap in the current FoV from the visual global map and rejects the outlier (points that are occluded or have depth discontinuity) in the submap. Then, a sparse-direct visual alignment is conducted to compute the frame-to-map image photometric errors. The LiDAR point-to-plane residual and the image photometric errors are tightly fused with the IMU propagation in an error-state iterated Kalman Filter. The fused pose is used to append new points to global map.
	
	\section{State estimation\label{estimation}}
	The state estimation of FAST-LIVO is a tightly-coupled error-state iterated Kalman filter (ESIKF) fusing measurements from LiDAR, camera and IMU. Here we mainly explain the system model (state transition model and measurement model). Readers can refer to \cite{he2021embedding} for the detailed structure and implementation of an on-manifold ISIKF.
	\vspace{-0.0cm}

		\begin{table}[t]
			\centering
			\caption{Some Important Notations}
			\label{tab:symbols}
			\begin{tabular}{lll}
				\toprule
				Notations  & Meaning \\
				\midrule
				${^G(\cdot)}$ & The vector ${(\cdot)}$ in global frame. \\
				${^C(\cdot)}$ & The vector ${(\cdot)}$ in camera frame.\\
				$^I\mathbf{T}_L$ & The extrinsic of LiDAR frame w.r.t. IMU frame. \\
				$^I\mathbf{T}_C$ &  The extrinsic of camera frame w.r.t. IMU frame. \\
				$\mathbf{x}, \hat{\mathbf{x}}$, $\bar{\mathbf{x}}$ & The ground-truth, predicted and updated estimation of $\mathbf{x}$.\\
				$\delta \mathbf{x}$ & The error state between ground-truth $\mathbf{x}$ and its estimation. \\
				\bottomrule
			\end{tabular}
			\vspace{-15pt}
		\end{table}
	\subsection{The boxplus ``$\boxplus$" and boxminus ``$\boxminus$'' operator} \label{boxplus}
	In this section, we use the ``$\boxplus$" and ``$\boxminus$" operations to express error of state on a manifold $\mathcal{M}$. Specifically, for $\mathcal{M} = SO(3)\times \mathbb{R}^n$ considered in this paper, we have:
	\begin{small}
		\begin{align}
			\begin{bmatrix}
				\mathbf{R} \\
				\mathbf{a}
			\end{bmatrix} 
			\boxplus 
			\begin{bmatrix}
				\mathbf{r} \\
				\mathbf{b}
			\end{bmatrix} 
			\triangleq \begin{bmatrix}
				\mathbf{R}\cdot\mathtt{Exp}(\mathbf{r}) \\
				\mathbf{a} + \mathbf{b}
			\end{bmatrix}, \hspace{0.1cm}
			\begin{bmatrix}
				\mathbf{R}_1 \\
				\mathbf{a}
			\end{bmatrix} 
			\boxminus
			\begin{bmatrix}
				\mathbf{R}_2 \\
				\mathbf{b}
			\end{bmatrix} 
			\triangleq \begin{bmatrix}
				\mathtt{Log}(\mathbf{R}_2 ^T \mathbf{R}_1 ) \\
				\mathbf{a} - \mathbf{b}
			\end{bmatrix}
			\nonumber
		\end{align}
	\end{small}
	where $\mathbf{r}\in \mathbb{R}^3$, $\mathbf{a}, \mathbf{b} \in \mathbb{R}^n$, $\mathtt{Exp}(\cdot)$ and $\mathtt{Log}(\cdot)$ represent the bidirectional mapping between the rotation matrix and rotation vector derived from the Rodrigues' formula\footnote[3]{\url{https://en.wikipedia.org/wiki/Rodrigues'\_rotation_formula}}.
	
	\subsection{State Transition Model}
	
	In our system, we assume the time offsets among the three sensors (LiDAR, IMU and camera) are known, which can be calibrated or synchronized in advance. We take IMU frame (donated as $I$) as the body frame and the first body frame as the gobal frame (donated as $G$). Besides, we assume that the three sensors are rigidly attached together and the extrinsics, defined in Table \ref{tab:symbols}, are pre-calibrated. Then, the discrete state transition model at the $i$-th IMU measurement:
	\begin{equation}
		\mathbf{x}_{i+1} = \mathbf{x}_{i} \boxplus \left(\Delta t\mathbf{f}\left(\mathbf{x}_i, \mathbf{u}_i, \mathbf{w}_i \right)\right)
		\label{eq_true_state_propagate}
	\end{equation}
	where $\Delta t$ is the IMU sample period, the state $\mathbf{x}$, input $\mathbf{u}$, process noise $\mathbf w$, and function $\mathbf{f}$ are defined as follows:
	\begin{align}
		\hspace{-0.2cm}
		\mathcal{M} &\triangleq SO(3) \times \mathbb{R}^{15},\ \text{dim}(\mathcal{M}) = 18  \nonumber \\
		\mathbf{x} &\triangleq
		\begin{bmatrix}
			^G\mathbf{R}_{I}^T & ^G\mathbf{p}_{I}^T & ^G\mathbf{v}^T & \mathbf{b}_{\mathbf{g}}^T & \mathbf{b}_{\mathbf{a}}^T & ^G\mathbf{g}^T
		\end{bmatrix}^T  \in \mathcal{M} \nonumber \\
		\mathbf{u} &\triangleq
		\begin{bmatrix}
			\boldsymbol{\omega}_{m}^T  & \mathbf{a}_{m}^T
		\end{bmatrix}^T, \hspace{0.2cm}
		\mathbf{w} \triangleq
		\begin{bmatrix}
			\mathbf{n}_{\mathbf{g}}^T  & \mathbf{n}_{\mathbf{a}}^T &
			\mathbf{n}_{\mathbf{b}\mathbf{g}}^T  & \mathbf{n}_{\mathbf{b}\mathbf{a}}^T
		\end{bmatrix}^T    \nonumber
	\end{align}
	\vspace{-0.6cm}
	\begin{small}
		\begin{align}
			\hspace{-0.0cm} \mathbf{f}(\mathbf{x}, \mathbf{u}, \mathbf{w} ) = 
			\begin{bmatrix}
				\boldsymbol{\omega}_{m} - \mathbf{b}_{\mathbf{g}} - \mathbf{n}_{\mathbf{g}}  \\
				^G\mathbf{v}+\displaystyle{\frac{1}{2}}(^G\mathbf{R}_{I}\left( \mathbf{a}_{m} - \mathbf{b}_{\mathbf{a}} - \mathbf{n}_{\mathbf{a}}\right)+{^{G}\mathbf{g}} )\Delta t \\
				^G\mathbf{R}_{I}\left( \mathbf{a}_{m} - \mathbf{b}_{\mathbf{a}} - \mathbf{n}_{\mathbf{a}}\right) + {^{G}\mathbf{g}}  \\
				\mathbf{n}_{\mathbf{b}_{\mathbf{g}}}\\
				\mathbf{n}_{\mathbf{b}_{\mathbf{a}}} \\
				\mathbf{0}_{3\times 1} 
			\end{bmatrix} \in \mathbb{R}^{18} \nonumber
		\end{align}
	\end{small}
	where $^G\mathbf{R}_I$ and $^G\mathbf{p}_I$ denote the IMU attitude and position in the global frame,  ${^G\mathbf{g}}$ is the gravity vector in the global frame, $\boldsymbol{\omega}_m$ and $\mathbf{a}_m$ are the raw IMU measurements, $\mathbf{n}_\mathbf{g}$ and $\mathbf{n}_\mathbf{a}$ are measurement noises in $\boldsymbol{\omega}_m$ and $\mathbf{a}_m$, $\mathbf{b}_\mathbf{a}$ and $\mathbf{b}_\mathbf{g}$ are IMU bias, which are modeled as random walk driven by Gaussian noise $\mathbf{n}_\mathbf{bg}$ and $\mathbf{n}_\mathbf{ba}$, respectively. 

	\subsection{Forward Propagation}\label{sec:propagation}
	We use the forward propagation to predict the state $ \hat{\mathbf{x}}_{i+1}$ and its covariance $\hat{\mathbf{P}}_{i+1}$, at each IMU input $\mathbf{u}_i$. More specifically, the state is propagated by setting the process noise $\mathbf{w}_i$ in (\ref{eq_true_state_propagate}) to zero: 
	\begin{align}
		\hat{\mathbf{x}}_{i+1} = \hat{\mathbf{x}}_{i} \boxplus \left( \Delta t \mathbf{f}(\hat{\mathbf{x}}_i, \mathbf{u}_i, \mathbf{0}) \right). 
		\label{eq:state_prop}
	\end{align}
	with covariance propagated as below:
		\begin{align}
			& \hat{\mathbf{P}}_{i+1} = \mathbf{F}_{\delta{\hat{\mathbf{x}}}} \hat{\mathbf{P}}_{i} \mathbf{F}_{\delta{\hat{\mathbf{x}}}}^T + \mathbf{F}_{\mathbf{w}} \mathbf{Q} \mathbf{F}_{\mathbf{w}}^T 
			\label{eq:cov_prop} \\
			&\mathbf{F}_{\delta{\hat{\mathbf{x}}}} = \left. \dfrac{ \partial  \delta{\hat{\mathbf{x}}}_{i+1}  }{\partial\delta{\hat{\mathbf{x}}_i}} \right|_{\delta{\hat{\mathbf{x}}_i} = \mathbf{0}, \mathbf{w}_i = \mathbf{0}} , ~~
			\mathbf{F}_{{\mathbf{w}}} = \left. \dfrac{ \partial \delta{\hat{\mathbf{x}}}_{i+1}   }{\partial{\mathbf{w}_i}} \right|_{\delta{\hat{\mathbf{x}}_i} = \mathbf{0}, \mathbf{w}_i = \mathbf{0}} \nonumber 
		\end{align}
	where $\mathbf{Q}$ is covariance of $\mathbf{w}$, $\delta \hat{\mathbf{x}}_{i} \triangleq \mathbf{x}_{i} \boxminus \hat{\mathbf{x}}_{i}$, and the concrete forms of $\mathbf{F}_{\delta{\hat{\mathbf{x}}}}$ and $\mathbf{F}_{{\mathbf{w}}}$ can be found in \cite{lin2021r2live, xu2020fastlio}. 
	
	The state prediction (\ref{eq:state_prop}) and covariance (\ref{eq:cov_prop}) propagate from time $t_{k-1}$, where the last LiDAR or image measurements are received, until time $t_k$, where the current LiDAR or image measurements are received, with the reception of each IMU measurements $\mathbf u_i$ in between $t_{k-1}$ and $t_k$. The initial state and covariance in (\ref{eq:state_prop}) and (\ref{eq:cov_prop}) are $\bar{\mathbf{x}}_{k-1}$ and $\bar{\mathbf{P}}_{k-1}$,which were obtained by fusing the last LiDAR or image measurement (see Section \ref{KF}).  We denote the state and covariance propagated until $t_k$ as $\hat{\mathbf{x}}_{k}$ and $\hat{\mathbf{P}}_{k}$, respectively. Notice that we do not assume LiDAR scans and images are received at the same time. The arrival of either a LiDAR scan or image will cause an update of the state as detailed in Section \ref{KF}. 
	\subsection{Frame-to-map Measurement Model}
	
	\subsubsection{LiDAR Measurement Model}\label{sec:lidar_meas_model}
	If a LiDAR scan is received at time $t_k$, we first conduct backward propagation proposed in \cite{xu2020fastlio} to compensate the motion distortion. The resultant points $\{{^{L}\mathbf{p}_j}\}$ in the scan can be viewed as being sampled simultaneously at $t_k$ and expressed in the same LiDAR local frame $L$. When registering the scan points  $\{{^{L}\mathbf{p}_j}\}$ to the map, we assume each point lies on a neighboring plane in the map with normal  $\mathbf{u}_j$ and center point ${{\mathbf{q}}_j }$. That is, if transforming the measured ${^{L}\mathbf{p}_j}$ expressed in the LiDAR local frame to the global frame using the ground-truth state (i.e., pose) $\mathbf x_k$, the residual should be zero: 
	
	\begin{small}
		\begin{align}\label{lidar_meas_model}
			\mathbf 0 = \mathbf{r}_l(\mathbf{x}_{k}, {^{L}\mathbf{p}}_{j}) = \mathbf{u}_j^T ({}^G{{\mathbf T}}_{I_k} {}^{I}{\mathbf T}_{L} \,^L{\mathbf p}_j - {{\mathbf{q}}_j } )
		\end{align}
	\end{small}
	
	In practice, in order to find the neighboring plane, we transform  ${^{L}\mathbf{p}_j}$ to the global frame using pose in the predictede state $\hat{\mathbf{x}}_{k}$ by
	$
	^G{\hat{\mathbf{p}}_j}={}^G{\hat{\mathbf T}}_{I_k} {}^{I}{\mathbf T}_{L} \,^L{\mathbf p}_j
	$ and search for the nearest 5 points in the LiDAR global map organized by an incremental kd-tree structure, \textit{ikd-tree} \cite{9697912}, to fit a plane. Then, the equation in (\ref{lidar_meas_model}) defines an implicit measurement model for state $\mathbf x_k$. In order to account for the measurements noise in $^L{\mathbf p}_j $, the equation is weighted by a factor $\boldsymbol{\Sigma}_l$.

	\begin{figure}[htp]
		\centering
		\includegraphics[width=0.7\linewidth]{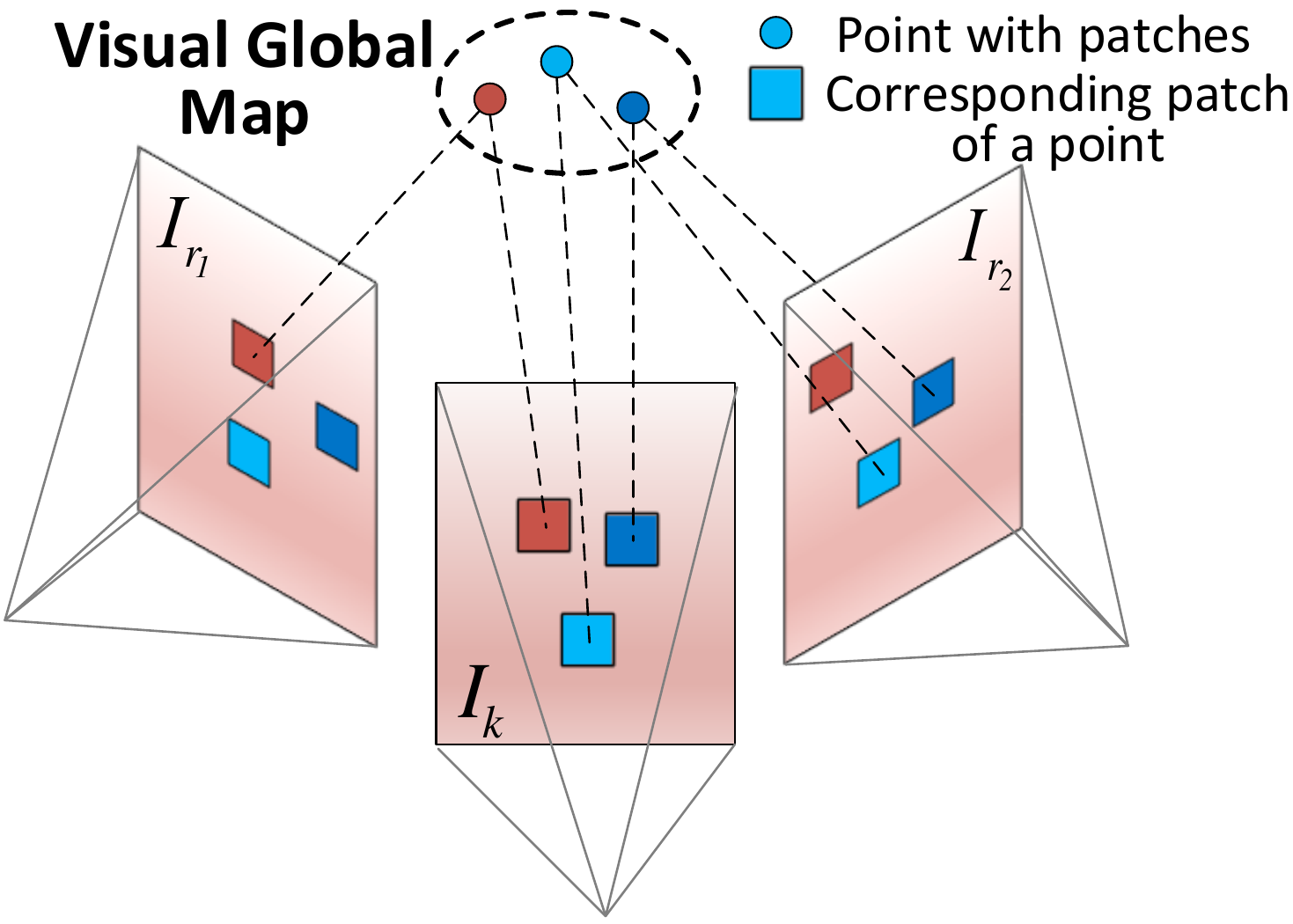}
		\caption{Changing the pose of current frame $I_k$ in the global frame to minimize the photometric errors between the reference image patches in the visual global map and  the corresponding patch in the current frame. }
		\label{fig:Frame_to_map}
		\vspace{-0.0cm}
	\end{figure}
	\subsubsection{Sparse-Direct Visual Alignment Measurement Model}\label{sec:image_align}
	
	Unlike the frame-to-frame image alignment in \cite{forster2014svo}, we conduct sparse-direct frame-to-map image alignment by minimizing the photometric errors in a coarse-to-fine manner, see Fig. \ref{fig:Frame_to_map}. Specifically, if an image is received at time $t_k$, we extract map points $\{^{G}{\mathbf{p}}_{i}\}$, from the global visual map, that fall within the image FoV (see Section \ref{submap}). For each map point ${^{G}{\mathbf{p}}}_{i}$, it has been attached with patches observed in different previous images (see Section \ref{Visual Global Map Data Structure}), we choose the path contained in the image that observes the point with the closest observation angle with the current image as the reference path (denoted as $\mathbf{Q}_i$). Then, transforming the map point ${^{G}{\mathbf{p}}}_{i}$ to the current image $\mathbf I_k(\cdot)$ with the ground-truth state (i.e., pose) $\mathbf x_k$, the photometric error between $\mathbf{Q}_i$ and respective path in the current image should be zero:
	
	\begin{small}
	\begin{equation} \label{e:visual_meas_model}
	    \begin{aligned}
			\mathbf 0 &=\mathbf{r}_ c ({{\mathbf{x}}_{k}, {^{G}{\mathbf{p}}}_{i}} ) =  \mathbf{I}_k(\boldsymbol{\pi}	(^{I}\mathbf{T}_C^{-1}\, ^{G} {\mathbf{T}}_{I_k}^{-1}{^{G}\mathbf{p}_i})) - \mathbf{A}_{i}\mathbf{Q}_i 
	\end{aligned}
	\end{equation}
	\end{small}
	where $\boldsymbol{\pi}(\cdot)$ is the pinhole projection model. 	The equation in (\ref{e:visual_meas_model}) defines another implicit measurement model for state $\mathbf x_k$ and is optimized (see Section \ref{KF}) at three levels, where on each level the current image and reference path is half sampled from the previous one.  The optimization starts from the coarsest level, after the convergence of a level, the optimization proceeds to the next finer level. In order to account for the measurements noise in the image $\mathbf I_k$, the equation is weighted by a factor $\boldsymbol{\Sigma}_c$.

	\subsection{Error-state Iterated Kalman Filter Update}\label{KF}

		\begin{table*}[]
			\centering
			\caption{Absolute translational errors (RMSE, METERS ) in NTU-VIRAL sequences with good quality ground truth}
			\label{tab:data}
			\begin{tabular}{llllllllll} 
				\toprule
				& eee\_01  & eee\_02 & eee\_03 & nya\_01 & nya\_02 & nya\_03 & sbs\_01  & sbs\_02 & sbs\_03\\ 
				\midrule
				FAST-LIVO  & \textbf{0.28} & \textbf{0.17} & \textbf{0.23} & \textbf{0.19} & \textbf{0.18} & \textbf{0.19} & 0.29 & \textbf{0.22} & \textbf{0.22}\\
				FAST-LIO2  & 0.54 & 0.22 & 0.25 & 0.24 & 0.21 & 0.23 & \textbf{0.25} & 0.26 & 0.24\\
				SVO2.0 (edgelets+prior)    & Fail & Fail & 4.12 & 2.29 & 2.91 & 3.32 & 7.84 & Fail & Fail\\
				R2LIVE & 0.45 & 0.21 & 0.97 & 0.19 & 0.63 & 0.31 & 0.56 & 0.24 & 0.44\\
				DVL-SLAM (no loop closure)   & 2.88 & 1.65 & 3.08 & 2.09 & 1.45 & 1.82 & 1.08 & 2.31 & 2.23\\
				\bottomrule
			\end{tabular}
		\end{table*}
		\vspace{-0.1cm}

	
	The propagated state $\hat{\mathbf{x}}_{k}$ and covariance $\hat{\mathbf{P}}_{k}$ derived from Section \ref{sec:propagation} impose the prior distribution for $\mathbf{x}_k$ as follows:
	\begin{equation}
		\label{eq:prior}
			\mathbf{x}_{k} \boxminus \hat{\mathbf{x}}_{k} 
			\sim \mathcal{N}(\mathbf{0}, \hat{\mathbf{P}}_{k}).
	\end{equation}

	Combining the prior distribution in (\ref{eq:prior}), the measurement distribution for LiDAR measurement in (\ref{lidar_meas_model}) and visual measurement in (\ref{e:visual_meas_model}), we obtain the maximum a posteriori (MAP) estimation for $\mathbf{x}_{k}$:

	\begin{small}
		\begin{align}
			\hspace{-0.5cm}\mathop{\min}_{{\mathbf{x}}_{k} \in \mathcal{M} } & \left(  \left\| {\mathbf x}_{k} \boxminus \hat{\mathbf{x}}_{k}  \right\|_{\hat{\mathbf{P}}_{k}}^2 + \sum\nolimits_{j=1}^{m_l} {  \| \mathbf{r}_l ({\mathbf{x}}_{k}, {^{L}{\mathbf{p}}}_{j})  \|^2_{\boldsymbol{\Sigma}_{l}} } \right. \nonumber \\
			& + \left. \sum\nolimits_{i=1}^{m_c}{ \| \mathbf{r}_c ({{\mathbf{x}}_{k}, {^{G}{\mathbf{p}}}_{i}}) \|^2_{\boldsymbol{\Sigma}_{c}} } \right)
			\label{eq_optimial_map}
		\end{align}
	\end{small}
	\hspace{-0.15cm}
	where $\left\| \mathbf{x} \right\|_{\boldsymbol{\Sigma}}^2 = \mathbf{x}^T \boldsymbol{\Sigma}^{-1} \mathbf{x}$. Notice that if a LiDAR scan is received at $t_k$, (\ref{eq_optimial_map}) fuses only LiDAR residual $\mathbf r_l$ with IMU propagation (i.e., $m_c = 0$). Similarly, if an image is received at $t_k$, (\ref{eq_optimial_map}) fuses only visual photometric error $\mathbf r_c$ with IMU propagation (i.e., $m_l = 0$).
	
	 The optimization in (\ref{eq_optimial_map}) is non-convex and can be iteratively solved by a Gauss-Newton method. Such an iterative optimization has been proven to be equivalent to an iterated Kalman filter \cite{bell1993iekf}. To deal with the manifold constraint $\mathcal{M}$, in each optimization iteration, we parameterize the state in the tangent space (i.e., the error state) of the current state estimate via the $\boxplus$ operation in Section \ref{boxplus}. The solved error state then updates the current state estimate and proceeds to the next iteration until convergence. The converged state estimate, denoted as $\bar{\mathbf x}_k$, and the Hessian matrix of (\ref{eq_optimial_map}) at convergence, denoted as $\bar{\mathbf P}_{k}$, are used to propagate the incoming IMU measurements as described in Section \ref{sec:propagation}.  The converged state is also used to update the new LiDAR scan to global map in Section \ref{Global Map for Direct LiDAR Odometry} and in Section \ref{Visual Global Map Data Structure} for visual global map.
	
	\section{Map management\label{map-management}}
	
	Our map consists of a point cloud map (the LiDAR global map) for the LIO subsystem and a point map attached with patches (the visual global map for the VIO subsytem. 
	
	\subsection{LiDAR Global Map\label{Global Map for Direct LiDAR Odometry}}
	
	Our LiDAR global map is adopted from FAST-LIO2 \cite{9697912}, which consists of all past 3D points organized into an incremental k-d tree structure \textit{ikd-Tree}\cite{cai2021ikd}. The \textit{ikd-Tree} provides interfaces of points inquiry, insertion, and delete. It also internally down samples the point cloud map at a given resolution, repeatedly monitors its tree structure, and dynamically balances the tree structure by rebuilding the respective sub-trees. When receiving a new LiDAR scan, we poll each point transformed with the predicted pose in the \textit{ikd-Tree} for nearest points (Section \ref{sec:lidar_meas_model}). After the scan is fused with IMU to obtain $\bar{\mathbf x}_k$ (Section \ref{KF}), we use it to transform the scan points to the global frame and insert them to the \textit{ikd-Tree} at the LIO rate. 
	
	\subsection{Visual Global Map\label{Visual Global Map Data Structure}} 
	
	The visual global map is a collection of LiDAR points previously observed. Each point is attached with multiple patches from the images observing it.  The data structure and update of the visual global map are explained below: 
	
	\subsubsection{Data Structure}
	
	To quickly find the visual map points falling in the current FoV, we use axis-aligned voxels to contain the points in visual global map. Voxels are of the same size and organized by a Hash table for fast indexing. A point contained in a voxel is saved with its position, multiple patch pyramids extracted from different reference images, and the camera pose of each patch pyramids. 
	
	\subsubsection{Visual Submap and Outlier Rejection\label{submap}}
	
	Even the number of voxels is much less than that of visual map points, determining which of them are within current frame FoV could still be very time consuming, especially when the map points (hence voxels) are large in number. To address this issue, we poll these voxels for each point of the most recent LiDAR scan. This can be done very efficiently by inquiring the voxel Hash table. If the camera FoV is roughly aligned with the LiDAR, map points falling in the camera FoV are mostly likely contained in these voxels. Hence, the visual submap can be obtained by the points contained in these voxels followed by a FoV check.
	
\begin{figure}[t]
		\centering
		\includegraphics[width=1.0\linewidth]{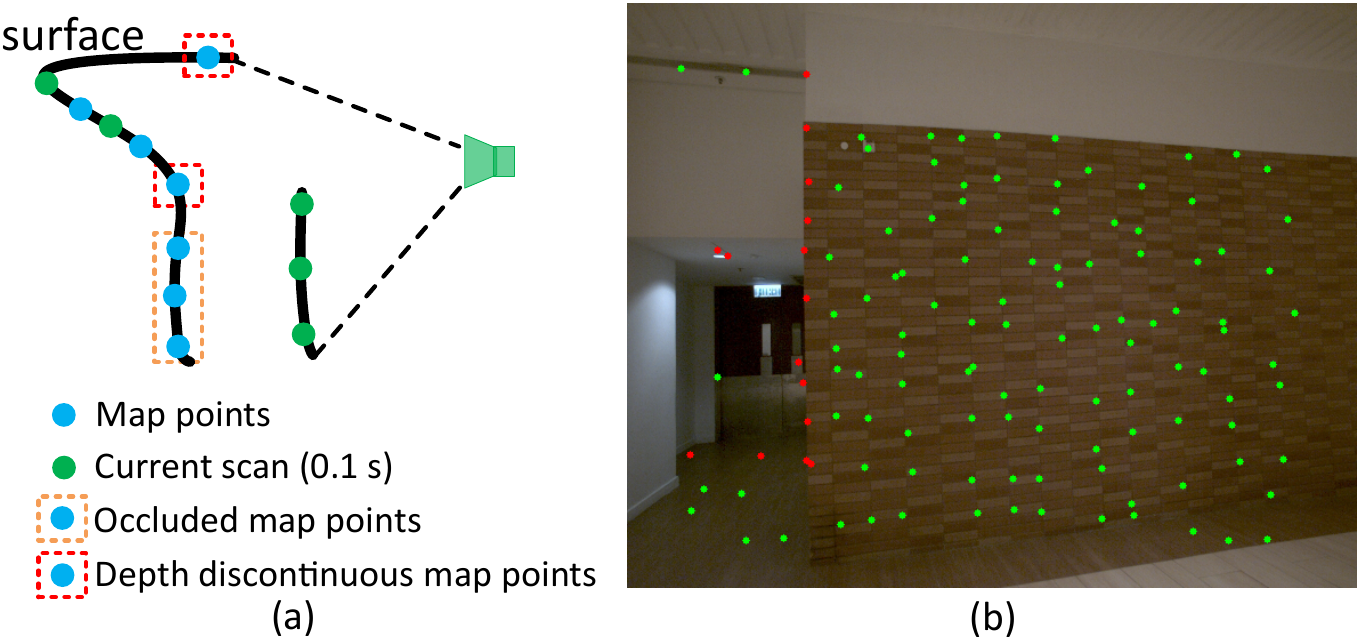}
		\caption{Outlier rejection. (a) shows the diagrammatic drawing of occluded and depth-discontinuous map points. (b) shows the effect visualization of outlier rejection module in real scenes. The red dots are the rejected map points, and the green dots are the accepted map points.}
		\label{fig:outlier}
	\end{figure}

	The visual submap could contain map points that are occluded in the current image frame or have discontinuous depth, which severely degrade the VIO accuracy.  To address this issue, we project all the points in the visual submap onto the current frame using the predicted pose in $\hat{\mathbf x}_k$ and keep the lowest-depth points in each grid of $40 \times 40$ pixels. Furthermore, we project the points in the most recent LiDAR scan  to the current frame and check if they occlude any map points projected within $9\times 9$ neighbor by examining their depth. Occluded map points are rejected (see Fig. \ref{fig:outlier}) and the rest will be used to align the current image (Section \ref{sec:image_align}). 
	
	\subsubsection{Update Visual Global Map}\label{sec: update_visual_map}
	
 	After a new image frame is aligned (Section \ref{sec:image_align}), we attach patches from the current image to map points within the FoV, so that the map points will likely have effective patches with  uniformly distributed viewing angles. Specifically, we select map points with high photometric errors after the frame alignment, if it has been more than 20 frames from the last time the map point is added with a patch or the map point in the current frame is more than 40 pixels away from its pixel position in the last reference frame where a patch was added, we add a new patch to it. The new patch is extracted from the current image with a size of $8\times8$ pixels. Along with the patch pyramid, we also attach the frame pose to the map point. 
	
	Besides adding patches to the map points, we also need to add new map points to the visual global map. To do so, we divide the current image into grids of $40\times40$ pixels and project on it the points in the most recent LiDAR scan. Projected LiDAR points with the highest gradient in each grid will be added to the visual global map, along with a patch extracted there and the image pose. To avoid LiDAR points on edges to be added to the visual map, we skip edge points with high local curvature \cite{zhang2014loam,lin2019loam_livox}. 

	\section{Experiment and Results}
	
	In this section, we verify our proposed method in both open and private datasets. 
	
	\subsection{Benchmark Dataset} 
	In this section, quantitative experiments are conducted on all the nine sequences of NTU-VIRAL open datasets \cite{nguyen2021ntuviral}. The sensors used in the dataset are the left camera, the horizontal 16-channel OS1 gen1\footnote[4]{\url{https://ouster.com/products/scanning-lidar/os1-sensor/}} LiDAR and its internal IMU. We compare our method with various open-source odometry system, including R2LIVE \cite{lin2021r2live}, a feature-based, full LiDAR-inertial-visual odometry, FAST-LIO2 \cite{9697912}, a direct, LiDAR-inertial odometry, SVO2.0 \cite{forster2016svo}, a semi-direct, visual-inertial odometry, and DVL-SLAM \cite{shin2020dvl}, a direct, LiDAR-visual SLAM system. All these systems are downloaded from their respective Github websites where the FAST-LIO2, R2LIVE, and DVL-SLAM all use their recommended parameters for outdoor scenarios. The parameters  of SVO2.0, including grid size, feature-selection threshold, and sliding window size, are tuned to to attain the best results to the authors' efforts. Moreover, since all systems in comparison are odometry with no loop closure, except DVL-SLAM, we remove the loop-closure module of DVL-SLAM to make a fair comparison. The modified DVL-SLAM, which we also opened on Github\footnote[5]{\url{https://github.com/xuankuzcr/DVL\_SLAM\_ROS}}, keeps the sliding window optimization to retain the accuracy as much as possible. 
	
	The results of all methods are shown in Table \ref{tab:data}, where each method uses the same parameters across all sequences. As can be seen, our method achieves the best accuracy among all sequences except the sequence ``sbs\_01", where our system has a slightly higher error than the LiDAR-inertial only odometry, FAST-LIO2. This is because the image is extremely blurred due to the fast UAV motion, fusing these low-quality images did not help on the odometry accuracy. Other than ``sbs\_01", our method benefits from the tight couple of all LiDAR, inertial, and visual information and its performance surpasses the LiDAR-inertial only odometry, FAST-LIO2.0 (by an extent depending on the image quality) and also other subsystems (e.g., visual-inertial only odometry, SVO2.0, and LiDAR-visual only odometry, DVL-SLAM) on all the rest sequences. Moreover, our system achieves higher accuracy than R2LIVE, a state-of-the-art feature-based tightly-coupled full LiDAR-inertial-visual odometry system. It is also interesting to see that R2LIVE occasionally achieves lower accuracy than the LiDAR-inertial only odometry, FAST-LIO2. The reason is due to the feature-based LIO subsystem of R2LIVE, which leads to fewer LiDAR points to be registered, while FAST-LIO2 has a direct LIO subsystem that registers more raw LiDAR points. The fusion of images in R2LIVE can compensate this loss of accuracy and therefore its performance surpasses FAST-LIO2 in some sequences.  Finally, it is noted that SVO2.0 fails in eee\_01, eee\_02, sbs\_02 and sbs\_03 due to the insufficient prior information caused by the fast UAV rotation and the strong image blur, while DVL-SLAM survived in these sequences due to the use of LiDAR measurements in its VO module.
	
	\subsection{Private Dataset}
	\subsubsection{Equipment Setup}\label{platform}
\begin{figure}[htp]
		\centering
		\includegraphics[width=1.0\linewidth]{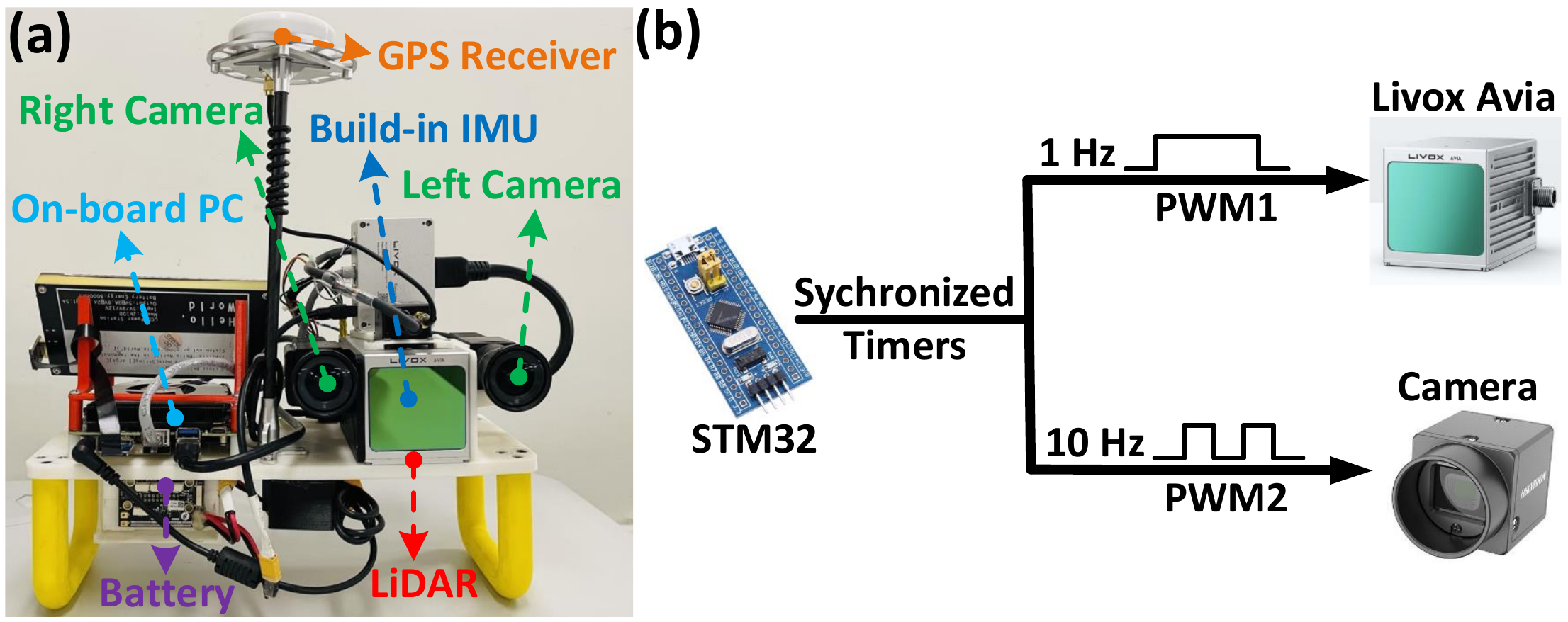}
		\caption{Our platform with hardware synchronization for data acquisition. (a) our hardware system, (b) the  hardware synchronization diagram.}
		\vspace{-0.0cm}
		\label{fig_handheld_device}
	\end{figure}

	Our sensor suite for data collection is shown in Fig. \ref{fig_handheld_device}, which consists of an onboard computer DJI manifold-2c (Intel i7-8550u CPU and 8 GB RAM), two industrial cameras (MV-CA013-21UC),  and a Livox Avia LiDAR. All sensors are hard synchronized with a trigger signal at 10 Hz generated by STM32 synchronized timers.

	\subsubsection{LiDAR Degenerated Experiment}
\begin{figure}[htp]
		\centering
		\includegraphics[width=1.0\linewidth]{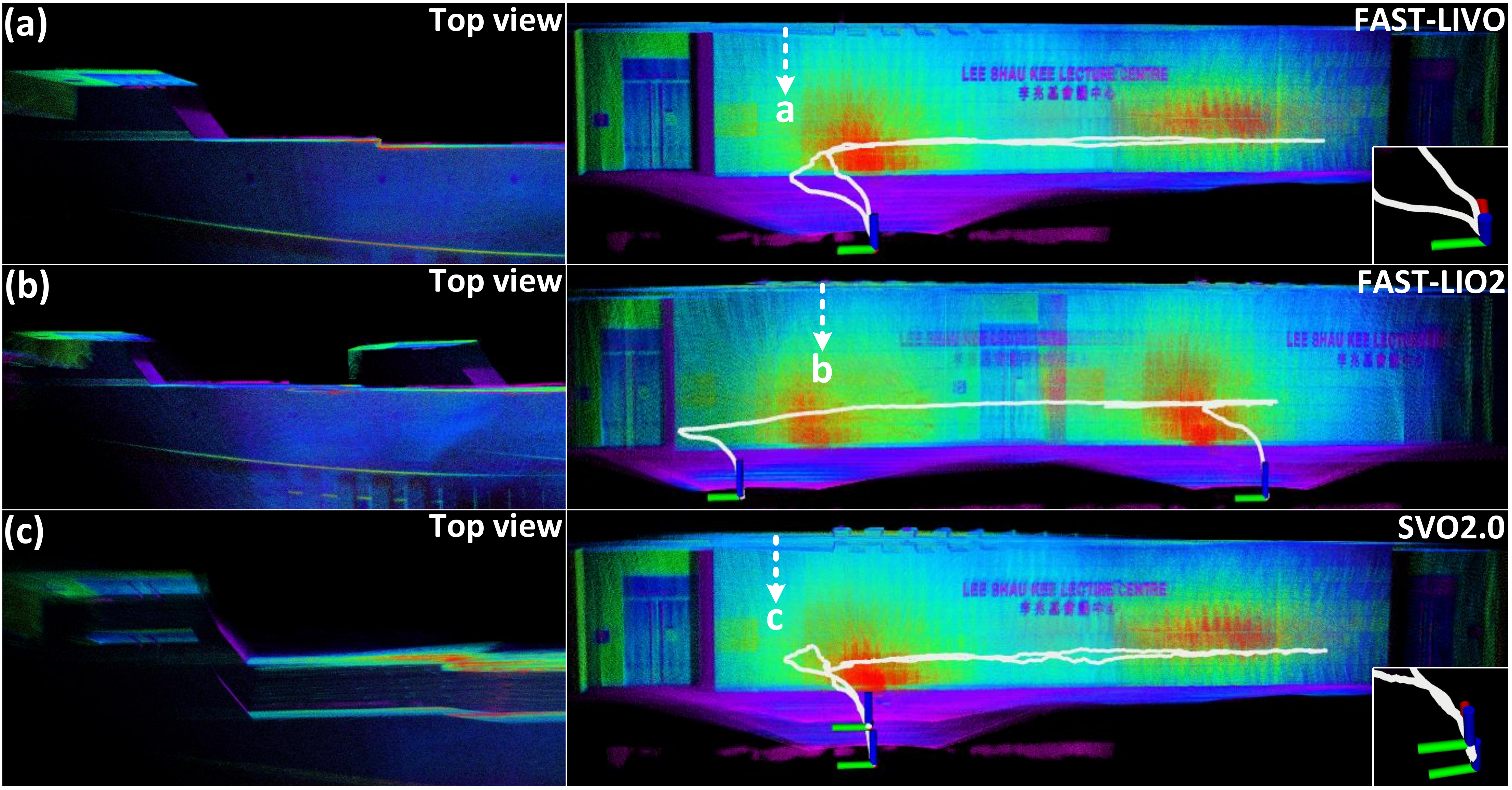}
		\caption{The performance of FAST-LIVO, FAST-LIO2, and SVO2.0 in a LiDAR degeneration scene. The point cloud is colored by LiDAR intensity. (a), (b) and (c) represent the top view of the corresponding position a, b, c respectively.}
		\label{LiDAR_Degenerated}
	\end{figure}

	In this experiment, we evaluate our system in a LiDAR degenerated environment, facing a wall with a length of about 30 meters. The results are shown in Fig. \ref{LiDAR_Degenerated}. Due to the lack of constraints in the direction along the wall, the LiDAR-inertial odometry, FAST-LIO2, cannot have a good state estimation in this direction, especially only a side wall can be seen in the FoV. Similarly, the highly repeated visual features on the wall and insensitive visual constraints in the vertical direction of the wall bring drifts to visual-inertial odometry, SVO2.0. Compared with FAST-LIO2 and SVO2.0, our system is robust and accurate to handle the scene and achieves the best performance, with the lowest end-to-end drift of 0.05m.
	\vspace{-2pt}
	\subsubsection{Visual Challenge Experiment}
\begin{figure}[htp]
		\centering
		\includegraphics[width=1.0\linewidth]{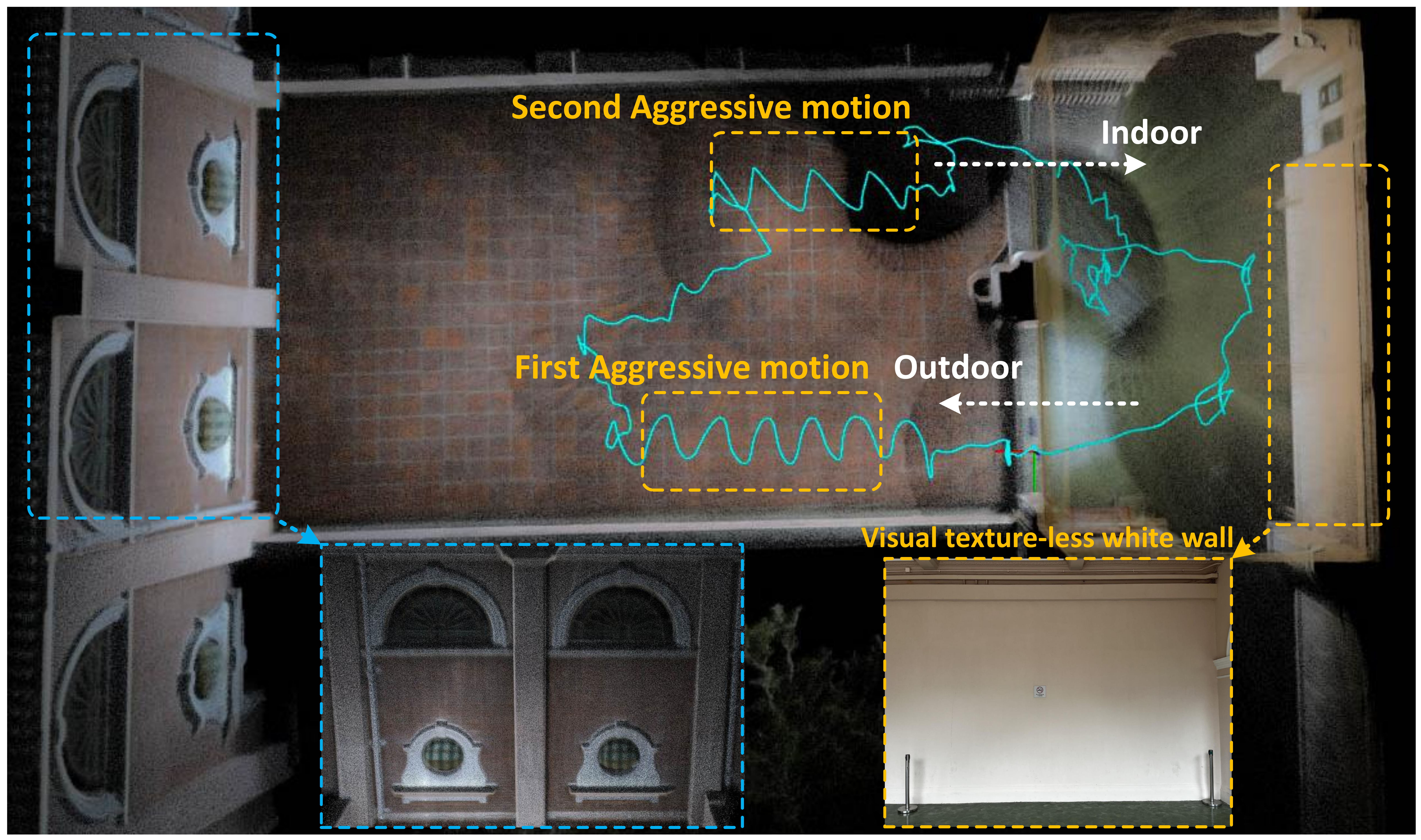}
		\caption{The RGB-colored point cloud of the visual challenging scene containing aggressive motions, indoor-to-outdoor and outdoor-to-indoor movements, and a visual texture-less wall. The green path is the computed trajectory.}
		\vspace{-2pt}
		\label{Visual_Degenerated}
	\end{figure}

	In this experiment, we challenge an extremely difficult scene in visual SLAM. As shown in Fig. \ref{Visual_Degenerated}, the motion of the sensor includes an indoor-to-outdoor process, an outdoor-to-indoor process, two aggressive motions and a visual texture-less white wall. Our proposed algorithm can reliably survives in this challenging scene and returns to the starting point with an end-to-end error of 0.04 m over the total path length 79.52 m.
	\subsubsection{High Precision Mapping with Colored Point Cloud}
\begin{figure}[htp]
		\centering
		\includegraphics[width=1.0\linewidth]{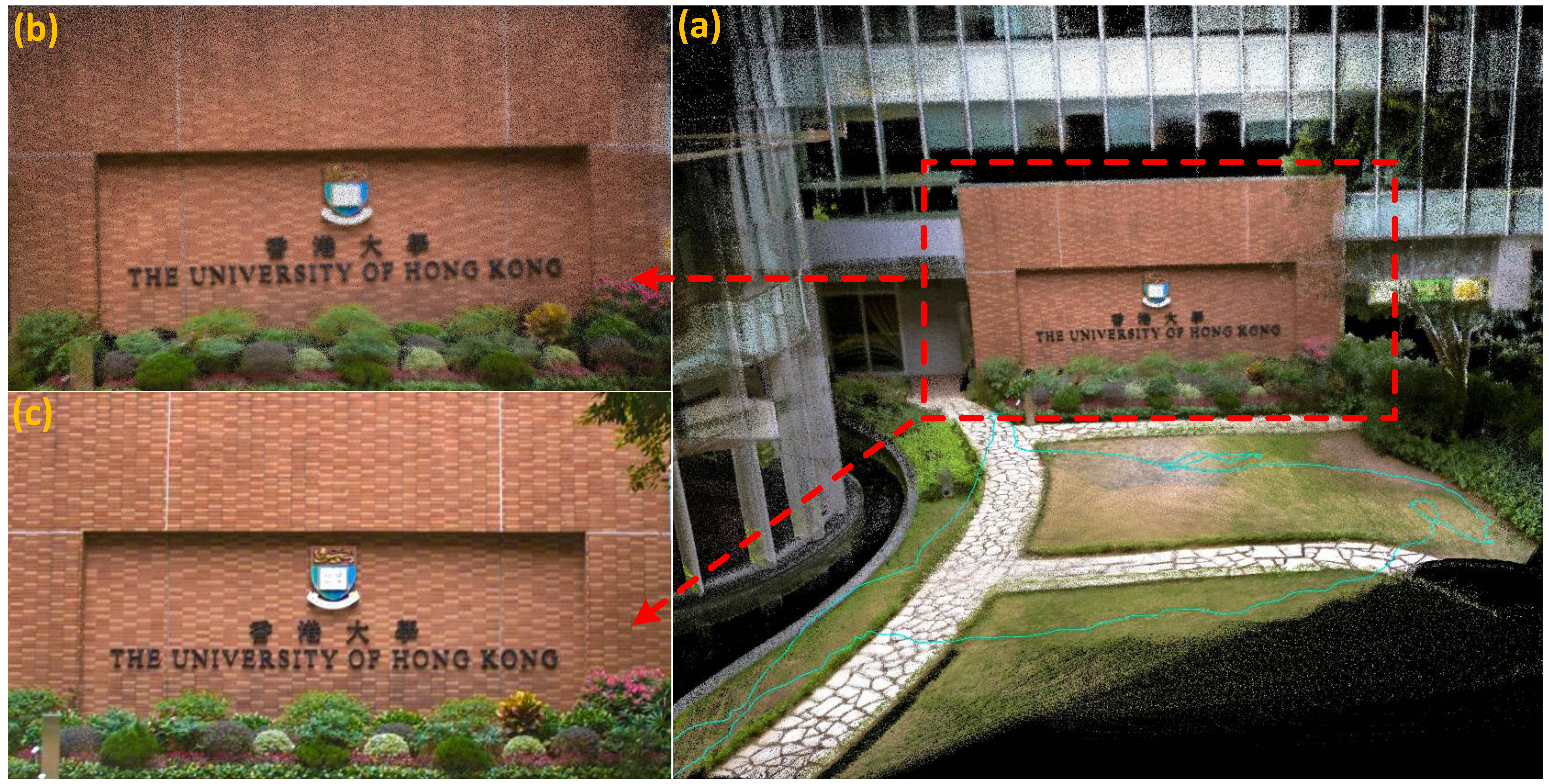}
		\caption{(a) RGB-colored point cloud, (b) partially enlarged view of the colored point cloud and (c) the corresponding RGB image.}
		\label{Colored}
	\end{figure}

	In this experiment, we take advantage of our algorithm
	to reconstruct, in real-time, a precise, dense, 3D, and RGB-colored map of a HKU campus environment, which is shown in Fig. \ref{Colored}. What worth to be mentioned
	is, the enlarged view of the colored point cloud has fine visual details similar to the actual RGB image.
	\vspace{-8pt}
	\vspace{-10pt}
		\begin{table}[hbp]
		\centering
		\caption{Mean time consumption in milliseconds}
		\label{tab:Time cosumption}
		\begin{tabular}{clll}
			\toprule
			\multicolumn{1}{l}{} &  & Intel i7 & ARM \\
			\midrule[0.6  
			pt]
			\multirow{6}{*}{VIO subsystem} & Visual Submap & 1.63 & 3.81 \\
			& Outlier Rejection & 2.07 & 2.58 \\
			& Sparse-Direct Visual Alignment & 3.19 & 3.88 \\
			& ESIKF Update & 0.42 & 0.52 \\
			& Update Visual Global Map & 2.92 & 3.03 \\
			& Total Time     & 10.23  & 13.82 \\
			\midrule[0.6pt]
			\multicolumn{2}{c}{LIO subsystem} & 26.52 & 51.51 \\		
			\bottomrule
		\end{tabular}
	\end{table}
	\vspace{-3pt}
	\subsection{Time Analysis}
	In this section, we evaluate the computation efficiency of our full system FAST-LIVO and compare it with a feature-based LiDAR-inertial-visual odometry system, R2LIVE. The comparison is performed among all private datasets on a desktop PC with a 8-core Intel Core i7-10700U processor, the mean time consumption per LiDAR and image frame of R2LIVE is 45.16 ms for the LIO and VIO frontend plus 59.27 ms for the VIO backend sliding window optimization, while FAST-LIVO is only 36.75 ms for the complete processing. Table \ref{tab:Time cosumption} further displays a mean time breakdown of FAST-LIVO on both the desktop PC and an embedded computation platform RB5 with a Qualcomm Kryo585 CPU. As can be seen, the FAST-LIVO can run in real-time on both the Intel and ARM processors with a significant computation margin. 
	
	\section{Conclusion}
	This paper proposes a fast, robust, sparse-direct, frame-to-map LiDAR-inertial-visual fusion framework, which is faster than the current state-of-the-art LIVO algorithms. We fuse the measurements of LiDAR, inertial, and camera sensors within an error-state iterated Kalman filter, utilizing the patch-based photometric error. Our system was tested in aggressive motions, indoor-outdoor, texture-less white wall, and LiDAR degenerated environments. The experiments in open datasets show our system can achieve the best overall performance among the state-of-the-art LIO, VIO, and LIVO algorithms.
	
	

	
	\section*{Acknowledgment}
	The authors gratefully acknowledge Livox Technology for the equipment support during the whole work. The authors would like to thank Jiarong Lin for the helps in the experiments.

	
	
	
	%
	
	\bibliographystyle{IEEEtran}
	\bibliography{paper}

\end{document}